\documentclass[10pt,twocolumn,letterpaper]{article}

\usepackage[pagenumbers]{cvpr} %
\usepackage{cuted}

\usepackage{algorithm}%
\usepackage{algpseudocode}%
\usepackage{array}
\usepackage{multirow}

\newcommand{\methodname}{\texttt{SRENDER}}
\definecolor{cvprblue}{rgb}{0.21,0.49,0.74}
\usepackage[pagebackref,breaklinks,colorlinks,allcolors=cvprblue]{hyperref}

\usepackage{amsmath,amsfonts,bm}

\def\eqref#1{equation~\ref{#1}}

\def\1{\bm{1}}

\definecolor{forward}{RGB}{117,33,33}

\def\rvx{{\mathbf{x}}}

\DeclareMathAlphabet{\mathsfit}{\encodingdefault}{\sfdefault}{m}{sl}
\SetMathAlphabet{\mathsfit}{bold}{\encodingdefault}{\sfdefault}{bx}{n}

\title{Efficient Camera-Controlled Video Generation of Static Scenes via Sparse Diffusion and 3D Rendering}

\author{Jieying Chen \quad Jeffrey Hu \quad Joan Lasenby \quad Ayush Tewari \\
University of Cambridge}

\begin{document}
\maketitle
\begin{strip}
    \centering
    \includegraphics[width=\linewidth]{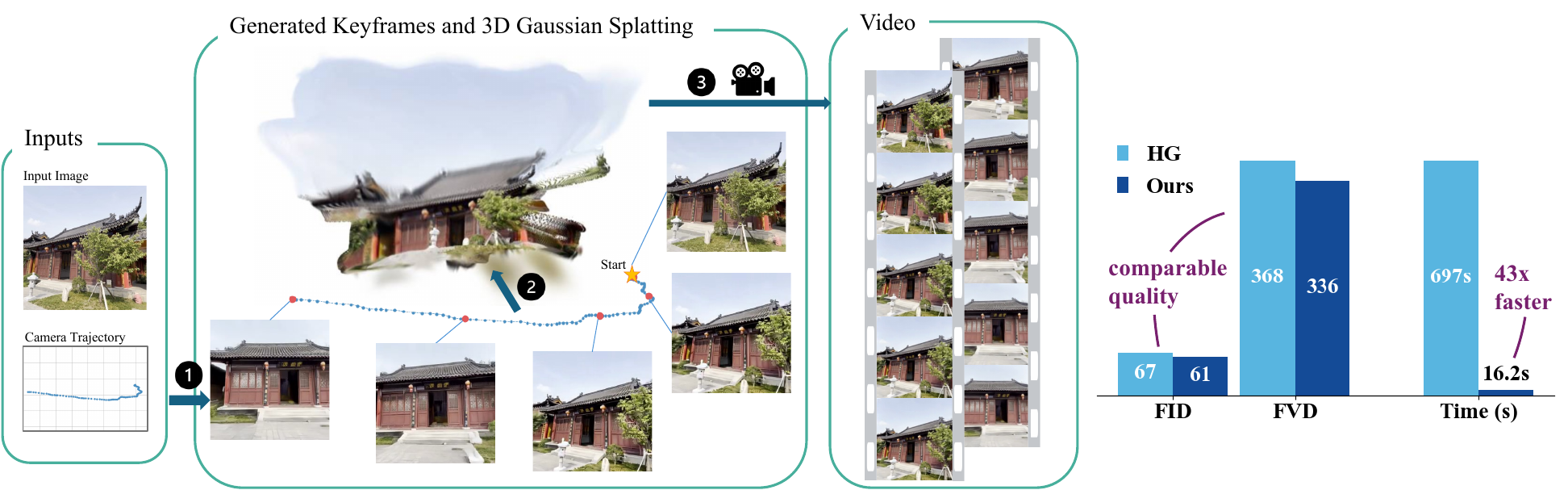}
    \captionof{figure}{\textbf{Teaser.} Left: Overview of our approach. Given an input image and a camera trajectory, 
    \methodname{} generates sparse keyframes, reconstructs the 3D scene, and renders the full video efficiently. Right: On average, our method is  roughly 43 times faster than the history-guided video diffusion baseline (HG)~\cite{song2025historyguidedvideodiffusion} when generating 20-second 30-fps videos from the DL3DV dataset, achieving real-time performance while maintaining comparable or better video quality.}
    \label{fig:teaser}
\end{strip}

\begin{abstract}
Modern video generative models based on diffusion models can produce very realistic clips, but they are computationally inefficient, often requiring minutes of GPU time for just a few seconds of video. This inefficiency poses a critical barrier to deploying generative video in applications 
that require real-time interactions, such as embodied AI and VR/AR.
This paper explores a new strategy for camera-conditioned video generation of static scenes: using diffusion-based generative models to generate a sparse set of keyframes, and then synthesizing the full video through 3D reconstruction and rendering. By lifting keyframes into a 3D representation and rendering intermediate views, our approach amortizes the generation cost across hundreds of frames while enforcing geometric consistency.
We further introduce a model that predicts the optimal number of keyframes for a given camera trajectory, allowing the system to adaptively allocate computation. Our final method, \methodname{}, 
uses very sparse keyframes for simple trajectories and denser ones for complex camera motion.
This results in video generation that is more than 40 times faster than the diffusion-based baseline in generating 20 seconds of video, while maintaining high visual fidelity and temporal stability, offering a practical path toward efficient and controllable video synthesis.
\end{abstract}

\section{Introduction}
\label{sec:intro}
Generative video models have recently achieved impressive visual fidelity. State-of-the-art models such as Sora \cite{Sora} and Wan \cite{wan2025} can synthesize photorealistic content that rivals cinematic footage. However, these successes come at a cost: current advances use diffusion-based~\cite{ho2020denoising} or flow-matching-based~\cite{lipman2022flow} techniques to generate every frame and are extremely computationally expensive. Producing even a short ten-second sequence can require up to tens of thousands of large neural network evaluations due to iterative denoising in diffusion models, amounting to several minutes of GPU time on high-end hardware. This inefficiency prevents real-time use and significantly limits applications in embodied AI applications, interactive content creation, and AR/VR. 

While some research has attempted to make these models more efficient by designing low-dimensional latent space  architectures~\cite{yu2024efficientvideodiffusionmodels, lin2025contentvefficienttrainingvideo}, or by utilizing distillation methods~\cite{yin2025causvid}, all existing methods still rely on neural networks to generate every frame of the video. 
In this paper, we challenge this paradigm. %
Videos are inherently redundant, as many frames depict the same underlying 3D scene under gradually varying conditions such as viewpoint, motion, or lighting.
Our goal is to develop a video generation framework that explicitly leverages these redundancies to enable efficient video synthesis.

In this work, we focus on the practically relevant setting of camera-conditioned generation of static scenes.
This allows us to study the core principle of leveraging scene redundancy for efficient video synthesis, without the additional complexity of object motion or deformation.
Camera-controlled video generation is an active area of research with several recent advances~\cite{song2025historyguidedvideodiffusion,wu2025geometryforcingmarryingvideo,huang2025voyagerlongrangeworldconsistentvideo}.
However, as previously mentioned, all these advances rely on generating every frame with neural networks. 
Recent work has demonstrated that incorporating 3D priors into video generation improves temporal consistency and novel-view controllability \cite{wu2025geometryforcingmarryingvideo, huang2025voyagerlongrangeworldconsistentvideo}. 
However, in all cases, 3D priors have only been used as an internal representation, or for auxiliary constraints. Every frame of the final video is still produced by diffusion-like neural generative models. 
Thus, while incorporating 3D priors has improved quality, it has not fundamentally addressed the efficiency barrier. This paper addresses this missing piece.

To this end, we present our method, \methodname{}, which generates full videos by first synthesizing a sparse set of keyframes with diffusion models and then generating the dense video through 3D scene reconstruction and rendering.
This approach leverages the inherent 3D structure of visual scenes: high-quality reconstructions can be obtained even from sparse multi-view observations, without requiring dense video input. 
The camera-controlled video can be rendered very efficiently from the reconstructed 3D scene using standard physically-based rendering techniques.
We build on advances in 3D reconstruction and rendering, particularly 3D Gaussian Splatting (3DGS)~\cite{kerbl20233dgaussiansplattingrealtime} that can render complex photo-realistic scenes at very high framerates, and follow-ups~\cite{jiang2025anysplatfeedforward3dgaussian,liu2025worldmirror} that can reconstruct 3DGS from multi-view observations with neural networks. 

\methodname{} also includes a model that adaptively selects how many keyframes to generate for a given camera trajectory. Simple trajectories with smooth motion or limited parallax can be reconstructed accurately from very few keyframes, whereas complex trajectories with large viewpoint changes require denser sampling. By predicting this keyframe budget directly from the camera path, \methodname{} allocates computation where it is most needed, maintaining visual fidelity while minimizing redundant generation.

Across videos up to twenty seconds long, our adaptive keyframe selection model chooses between 4 and 35 keyframes, corresponding to generating at most one-tenth as many frames as would be required by a standard 30 fps video.
Our approach achieves more than 40× speed-up on DL3DV~\cite{ling2023dl3dv10klargescalescenedataset} and more than 20× speed-up on RealEstate10k~\cite{realestate10k}, while maintaining comparable visual quality and temporal consistency.
These results demonstrate that explicit 3D reasoning and adaptive sparse generation can dramatically reduce the computational cost of video synthesis without sacrificing fidelity.

\section{Related Work}
\label{sec:related}

\subsection{Camera-controlled Video Models}

Video generative models have seen fast progress in recent years, driven by diffusion models~\cite{ho2020denoising}.
Most models focus on text-to-video, or image-to-video tasks~\cite{blattmann2023stablevideodiffusionscaling, wan2025, agarwal2025cosmos, veo3, Sora}. 
Following the success of latent diffusion models in the image domain, early video models trained 3D UNets in the latent space of 3D VAEs~\cite{blattmann2023stablevideodiffusionscaling}. Subsequent models replaced the UNet with a transformer for its greater scalability~\cite{wan2025, agarwal2025cosmos}. Most of the video diffusion models follow the standard diffusion procedure, denoising all frames of the video jointly from the same noise level.  A recent influential work, diffusion forcing, breaks from this trend by proposing to add independent noise levels to each frame~\cite{chen2024diffusionforcingnexttokenprediction}, and diffuse frames with different noise levels together. This strategy enables different denoising strategies at inference time, including autoregressive decoding for long-range video generation with variable-length condition frames, as well as video interpolation between keyframes. This piece of work then inspired a flurry of work on autoregressive video models~\cite{huang2025selfforcingbridgingtraintest, cui2025selfforcingminutescalehighqualityvideo, ai2025magi1autoregressivevideogeneration, chen2025skyreelsv2infinitelengthfilmgenerative, deng2025autoregressivevideogenerationvector, MirageLSD, yuan2025lumos1autoregressivevideogeneration, liu2025rollingforcingautoregressivelong}. Camera-controlled video generation is a subfield of video generation that has been shown to work very well. 
Existing methods for this problem either directly encode the desired camera poses to condition the generative model~\cite{he2024cameractrl, zhou2025stablevirtualcameragenerative}, or make explicit use of the 3D structure of the world, e.g., by computing a point cloud from input images and rendering it from the desired camera view to condition the generative model~\cite{ren2025gen3c3dinformedworldconsistentvideo, huang2025voyagerlongrangeworldconsistentvideo}. In this work, we focus specifically on the camera-controlled video generation problem, and we opt for a diffusion-forcing architecture. This design choice enables the generation of sparse keyframes with strong scene consistency and flexible conditioning on previous frames.

In general, inference with video diffusion models is very expensive. Even generating short (\textless 3s) clips can take minutes on high-end hardware~\cite{wan2025, hong2022cogvideo, yang2025cogvideoxtexttovideodiffusionmodels, ai2025magi1autoregressivevideogeneration}. 
Several attempts have been made at making these models faster, e.g., by using teacher-student distillation~\cite{yin2025causvid, cui2025selfforcingminutescalehighqualityvideo} or caching~\cite{yin2025causvid, lin2024animatedifflightningcrossmodeldiffusiondistillation, li2024t2vturbobreakingqualitybottleneck, wang2023videolcmvideolatentconsistency, zou2025acceleratingdiffusiontransformerstokenwise, huang2025selfforcingbridgingtraintest}.
However, all approaches rely on neural networks to generate every frame of the video, and do not take advantage of the information redundancies in the video signal. 
Our method is complementary to these advances. Any improvement in diffusion inference directly reduces the cost of generating sparse keyframes, while the overall efficiency gains of our method stem from not generating intermediate frames with neural networks at all.

\subsection{3D Reconstruction}
Advances in 3D representations, particularly 3D Gaussian Splatting (3DGS)~\cite{kerbl20233dgaussiansplattingrealtime}, have enabled high-quality reconstruction of scenes from collections of images. Many recent models train neural networks to regress 3DGS representations directly from posed input images~\cite{charatan2024pixelsplat3dgaussiansplats,tang2024dreamgaussiangenerativegaussiansplatting}. These approaches can recover detailed and consistent 3D geometry without any test-time optimization. More recent systems~\cite{jiang2025anysplatfeedforward3dgaussian, wang2025vggt} extend this capability by leveraging architectures inspired by DUST3R~\cite{wang2024dust3rgeometric3dvision}, enabling efficient 3D reconstruction even from unposed image sets~\cite{jiang2025anysplatfeedforward3dgaussian,liu2025worldmirror}.

Most progress in this area has focused on \emph{deterministic} reconstruction. Such models reconstruct only the parts of the scene directly visible in the input images and cannot represent the full distribution of plausible 3D scenes consistent with the observations. As a result, although they produce high-quality geometry, they cannot be used as generative models.
Some works have explored \emph{generative} 3D reconstruction. Early approaches~\cite{liu2023zero1to3zeroshotimage3d,gao2024cat3d} optimize a 3D representation using image diffusion models, but these optimization-based procedures are slow and typically limited to object-level scenes. More recent methods integrate diffusion and 3D representations~\cite{tewari2023forwarddiffusion,szymanowicz2023viewset}, enabling partially feed-forward 3D generation. However, these approaches have not yet achieved the visual quality, stability, or multi-view consistency seen in state-of-the-art video diffusion models.
Closely related are methods that first use image diffusion models to generate multi-view images from a single input image and then fit a 3D representation~\cite{liu2023syncdreamer,long2024wonder3d}. While these methods can produce coherent 3D assets, they remain restricted to object-centric settings and do not scale to full scenes or long camera trajectories. Recently, there have also been works that combine the generative capability of video diffusion models with an explicit 3D scene representation~\cite{huang2025voyagerlongrangeworldconsistentvideo, worldlabs2025marble, yu2024wonderworld}, achieving impressive results for generative 3D scene reconstruction.

Our method replaces dense video frame generation in the video model with a deterministic feed-forward 3D reconstruction model that reconstructs a 3DGS representation from the generated sparse keyframes. The video generation speed can thus be greatly improved as the deterministic feed-forward 3D reconstruction and rendering of the intermediate frames are much faster than diffusion-based frame synthesis.

\begin{figure*}[t]
    \centering
    \includegraphics[width=\linewidth]{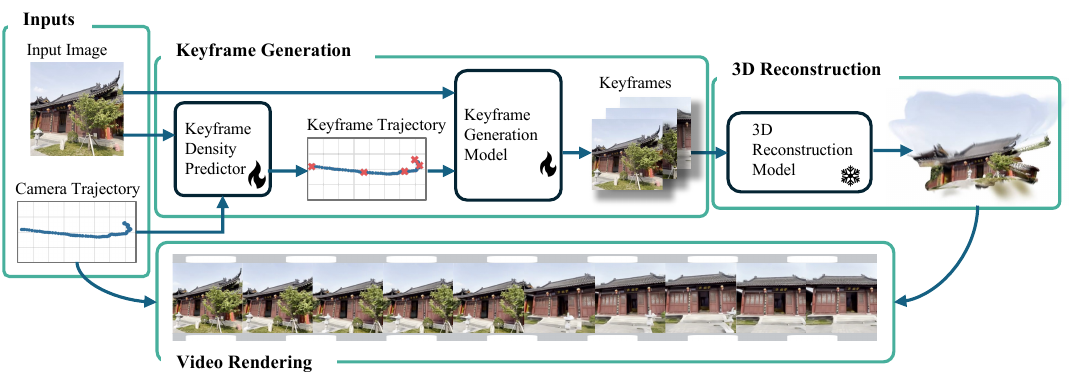}
    \vspace{-0.7cm}
    \caption{\textbf{Overview of \methodname{}.} Given an image and a target camera trajectory, a keyframe density predictor predicts the optimal keyframe density for the depicted scene and camera trajectory. Keyframe poses are then uniformly sampled along the trajectory before being fed to the keyframe generation model together with the input image for generating keyframes. A 3D reconstruction model takes the keyframes and generates the 3D representation of the scene. Finally, the video is rendered from the 3D scene along the input camera trajectory.}
    \label{fig:method}
\end{figure*}

\section{Keyframe Diffusion and 3D Rendering}
\label{sec:method}

Figure~\ref{fig:method} provides an overview of \methodname{}. Given an input image and a specified camera trajectory, the goal is to generate a video sequence that starts from the input image and corresponds to that trajectory. Rather than generating every frame through diffusion, \methodname{} synthesizes a sparse set of \emph{keyframes} and renders the remaining frames efficiently via 3D reconstruction.

We begin by using a \emph{keyframe density predictor} that analyzes the camera trajectory and determines the optimal sparsity of keyframes. This model adaptively allocates computation based on motion complexity.
Next, a diffusion-based \emph{keyframe generator} synthesizes the selected keyframes conditioned on the input image and camera poses corresponding to the keyframes. The resulting multi-view set captures the appearance of the static scene from diverse viewpoints along the conditioned camera trajectory. We then reconstruct a 3D Gaussian representation from these keyframes using a deterministic (non-generative) neural network. Once reconstructed, a dense and geometrically consistent video can be rendered at high frame rates.

\subsection{Adaptive Keyframe Selection}
\label{sec:keyframe_selection}

A central design choice in \methodname{} is determining how many keyframes to generate for a given camera trajectory. Dense sampling increases computational cost, while overly sparse sampling leads to incomplete 3D reconstructions and visible holes in rendered views. 

The optimal keyframe set depends jointly on the camera path and the underlying scene geometry. Smooth trajectories or limited parallax may require only a few keyframes, whereas large viewpoint changes or complex geometry demand denser sampling. We formulate this as a learning problem in which the optimal number of keyframes is predicted from the camera trajectory and scene appearance.

\paragraph{Model.}
We train a transformer-based keyframe density predictor. The model takes the full sequence of camera poses as input, each represented as an independent token. 
Scene appearance is incorporated by extracting the global feature token with a DINOv2~\cite{oquab2024dinov2learningrobustvisual} image encoder and appending it as an additional token. The tokens are processed by multiple self-attention blocks,
and the resulting features are averaged and passed through a lightweight MLP to predict the optimal number of keyframes. %

\paragraph{Supervision.}
\begin{algorithm}[t]
\caption{Keyframe Selection}
\label{alg:frame_selection}
\begin{algorithmic}[1]
\Require Frames $\mathcal{F} = \{F_1, F_2, \dots, F_N\}$, 
camera poses $\mathcal{P} = \{P_1, P_2, \dots, P_N\}$, 
coverage threshold $\tau$

\State \textbf{Initialization:}
\State \textbf{Point Cloud Generation:}
\State Obtain global point cloud $C_{global} = \text{VGGT}(\mathcal{F})$, consisting of sub-point clouds $\{C_1, C_2, \dots, C_N\}$ corresponding to each frame.
\State Initialize selected frame set $\mathcal{S} \gets \{F_1\}$ and combined point cloud $C \gets C_1$

\Statex
\State \textbf{Iterative Selection:}
\For{each subsequent frame $F_i \in \mathcal{F} \setminus \mathcal{S}$}
    \State Project current point cloud $C$ onto image plane using pose $P_i$
    \State Compute coverage ratio $r_i$ of projected points on $F_i$
    \If{$r_i < \tau$}
        \State $\mathcal{S} \gets \mathcal{S} \cup \{F_i\}$
        \State $C \gets C \cup C_i$
    \EndIf
\EndFor

\State \Return selected frame set $\mathcal{S}$
\end{algorithmic}
\end{algorithm}

Ground-truth keyframe densities are derived automatically from the RealEstate10k~\cite{realestate10k} and DL3DV~\cite{ling2023dl3dv10klargescalescenedataset} datasets. For each video, we reconstruct a point cloud using VGGT~\cite{wang2025vggt}. Starting from the first frame, we iteratively project the collective point cloud of the selected keyframes onto subsequent frames. When the projected coverage falls below a threshold, we mark the current frame as a new keyframe. We provide a pseudo-algorithm in Algorithm~\ref{alg:frame_selection}. This approach ensures that the selected keyframes collectively cover all pixels across the video. The transformer is trained to regress the number of keyframes produced by this procedure.

\paragraph{Discussion.}
This adaptive mechanism allocates generation effort where it is most beneficial: sparse keyframes are generated for smooth or low-parallax trajectories, and denser
ones for complex motion.
After the keyframe count is obtained, the keyframe camera poses are sampled uniformly along the camera trajectory.

\subsection{Keyframe Diffusion}

Once the keyframe density predictor selects the target poses, we synthesize the corresponding keyframes using a diffusion-based generator. The goal is to produce a small set of high-quality, geometrically consistent views that capture the static scene from the specified viewpoints.

Our keyframe generator builds on recent advances in camera-conditioned video diffusion, particularly diffusion forcing~\cite{chen2024diffusion} and history-guided video diffusion~\cite{song2025historyguidedvideodiffusion}, which improve temporal and geometric coherence across frames.

\subsubsection{Preliminaries}

In a standard denoising diffusion model~\cite{ho2020denoising}, a data sample $\rvx_0$ is gradually perturbed through a forward noising process $q(\rvx_t \mid \rvx_{t-1})$, and a neural network learns to reverse this process by predicting the noise at each step. Generation begins by sampling $\rvx_T \sim \mathcal{N}(0, \mathbf{I})$ and applying
\[
\rvx_{t-1} = f_\theta(\rvx_t, t, c),
\]
where $c$ denotes conditioning information such as camera pose.
In practice, the denoising is performed by a neural network with parameters $\theta$.

\paragraph{Diffusion Forcing and History Guidance.}
In video diffusion, $\rvx_0$ is a video composed of multiple image frames. Each frame of the output video needs to be denoised. 
Diffusion Forcing~\cite{chen2024diffusion} assigns an independent noise level to each frame, allowing the model to denoise frames with different noise levels together, thus possible to mix clean history frames with partially noised to-be-generated ones at arbitrary frame positions. 
Building on this, History-Guided Video Diffusion~\cite{song2025historyguidedvideodiffusion} applies classifier-free guidance~\cite{ho2022classifier} on subsets of frames (e.g., previously denoised frames) to encourage long-range consistency.

\subsubsection{Our Model}
\label{sec:keyframe diffusion method}

We treat the selected keyframes as a very low-frame-rate video and train a history-guided diffusion model to capture their joint distribution. Conditioning is provided by the first input frame and the camera trajectory. The first frame serves as a stable appearance anchor for all generated keyframes.

However, directly training a diffusion model at extremely low frame rates is unstable: the large viewpoint jumps cause geometric and photometric drift. To address this, we adopt a \emph{progressive training strategy}. We first train the model on high-frame-rate videos with dense supervision, then gradually reduce the effective frame rate by subsampling frames until it matches the sparse keyframe spacing used at inference. This process enables the model to learn short-range correspondences before handling large viewpoint changes, yielding stable and coherent multi-view generation even at high sparsity. 

The history-guided diffusion model uses a context window of 8 frames. To generate more than 8 consistent keyframes, we employ a two-stage inference scheme. First, we generate 8 keyframes that uniformly span the entire trajectory using only the provided input image as conditioning. 
Then, we generate the remaining keyframes using the same model, with the nearest already generated keyframes as conditioning.
This ensures coherence across arbitrarily long keyframe sequences while keeping the diffusion cost manageable.

\subsection{3D Reconstruction and Rendering}

Given the generated keyframes, we reconstruct a 3D representation of the static scene and render the dense video along the trajectory. We use the pretrained AnySplat model~\cite{jiang2025anysplatfeedforward3dgaussian}, which predicts a 3D Gaussian Splatting (3DGS) representation directly from a small set of unposed images.

\subsubsection{Preliminaries}
\paragraph{3D Gaussian Splatting.} 
3DGS~\cite{kerbl20233dgaussiansplattingrealtime} represents a scene using anisotropic Gaussian primitives parameterized by their mean, covariance, color, and opacity. Rendering is performed via differentiable rasterization in screen space. 3DGS offers real-time rendering performance while maintaining high visual fidelity.

\paragraph{AnySplat Reconstruction.}
AnySplat predicts Gaussian parameters from multiple unposed images in a single forward pass. Instead of iterative inverse rendering, it maps multi-view features to a 3D Gaussian field, enabling fast and reliable sparse-view reconstruction. It uses VGGT~\cite{wang2025vggt} to estimate camera poses, permitting training on unposed datasets. AnySplat is deterministic and reconstructs only the visible scene content. It processes tens of images in a few seconds.

\subsubsection{Our Model}
We feed the generated keyframes into AnySplat to obtain a 3D Gaussian representation of the scene. Since AnySplat’s predicted poses and the input camera trajectory lie in different coordinate frames, we align them by estimating a least-squares affine transformation.  
Finally, we render the dense output video by evaluating the 3DGS renderer at each camera pose. This stage is extremely fast and accounts for much of \methodname{}’s efficiency relative to diffusion-based baselines.

\subsection{Temporal Chunks}
\label{sec:temporal_chunks}

Our method relies on the generated keyframes being geometrically and photometrically consistent across the full camera trajectory. For simpler datasets such as RealEstate10k, and for short durations in more complex scenes, this assumption generally holds: diffusion models produce keyframes stable enough for a single high-quality global reconstruction.

However, on more challenging datasets such as DL3DV, we observe noticeable drift in the generated keyframes for long trajectories, using our model or other baseline models. Keyframes beyond roughly 10 seconds often exhibit inconsistencies in structure, appearance, or relative geometry. This reflects a broader limitation of current diffusion models, which struggle with long-range loop-closure and maintaining multi-view consistency across large-baseline changes.
Attempting to reconstruct a single global 3D scene from all keyframes leads to blurred reconstructions, as the 3D model must reconcile mutually inconsistent observations. To address this, we divide the keyframes into fixed-length \emph{temporal chunks}. In practice, we use chunk durations of 10 seconds, within which the generated keyframes are empirically consistent.

For each chunk, we perform an independent 3D reconstruction using AnySplat and align the dense input camera trajectory individually. Since each chunk produces its own 3D Gaussian representation in its own coordinate frame, we include a shared keyframe between adjacent chunks and align the scenes by estimating an affine transformation 
between the shared pose sets. After alignment, we render the dense video by querying the appropriate chunk-specific 3D model at each camera pose.
This chunked reconstruction strategy allows our method to generate high-quality videos without requiring the diffusion model to generate drift-free keyframes.

\section{Experiments}
\label{sec:experiments}
\begin{figure*}[t]
    \centering
    \includegraphics[width=0.8\linewidth]{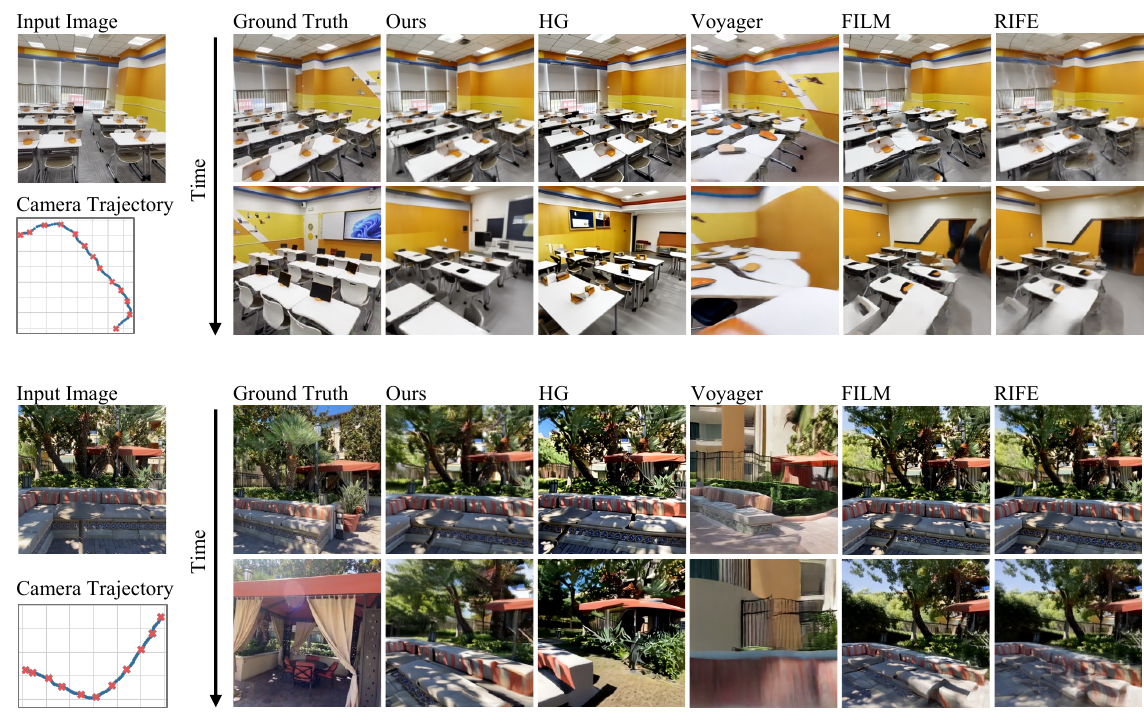}
    \caption{\textbf{Qualitative comparisons on DL3DV.} All methods generate 20-second videos at 5 fps, conditioned on the input image and target trajectory. Output frames at 4s and 14s are visualized. Our method achieves both high video quality and camera control. HG~\cite{song2025historyguidedvideodiffusion} often has high-frequency artifacts. Voyager~\cite{huang2025voyagerlongrangeworldconsistentvideo} fails at generating a consistent long video sequence. 
    We also show results with 2D interpolation methods~\cite{huang2022rife,reda2022film}, which show strong morphing effects, and also cannot satisfy the intermediate camera control inputs. }
    \label{fig:comparison_dl3dv}
\end{figure*}

\subsection{Experimental Details}

\paragraph{Datasets.}
We conduct experiments on two camera-conditioned video datasets: RealEstate10k (RE10K)~\cite{realestate10k} and DL3DV~\cite{ling2023dl3dv10klargescalescenedataset}. 
For RE10K, we generate 20-second videos at 10~fps (200 frames) along the provided camera paths. 
For DL3DV, which contains significantly larger camera motions, we generate 20-second videos at 30~fps (600 frames). 
We train separate models for each dataset, and train all models at a resolution of $256\times256$. 
For evaluation, we use 50 videos from the DL3DV test set and 200 videos from the RE10K test split.
Outside of the primary evaluation in Table~\ref{tab:comparison_combined}, 
we use a subsampled 5~fps version of DL3DV, 
as generating high-frame-rate sequences with the baseline methods is either very slow or not possible.
This test set allows us to perform comparisons on long sequences without high computational costs.

\paragraph{Video Generation Baselines.}
Our primary baseline is the History-Guided Video Diffusion model (HG)~\cite{song2025historyguidedvideodiffusion}. 
This baseline is particularly relevant because it shares the same diffusion architecture and training setup as our keyframe generator; the only difference is that HG generates every frame, whereas the keyframe generator in \methodname{} generates only sparse keyframes. 
We train HG on the same training splits as our model.

In addition, we compare against the state-of-the-art camera-conditioned model \emph{Voyager} from Hunyuan~\cite{huang2025voyagerlongrangeworldconsistentvideo}. 
Voyager is a very recent model that benchmarks against and improves upon multiple modern video generation approaches~\cite{gen-3,yang2025cogvideoxtexttovideodiffusionmodels, zhou2024allegroopenblackbox}, making it a strong representative baseline. We use the pretrained model of Voyager. The currently available implementation of Voyager is not capable of generating videos with hundreds of frames; therefore, we only compare it in the 5 fps setting.

\paragraph{2D Interpolation Baselines.}
To evaluate the importance of 3D reconstruction, we compare against two 2D frame interpolation methods: FILM~\cite{reda2022film} and RIFE~\cite{huang2022rife}. 
These methods interpolate intermediate frames between our generated keyframes, allowing us to assess whether 3D rendering offers advantages over purely 2D temporal interpolation.

\subsection{Evaluation Metrics}

\paragraph{Image and Video Quality.}
We evaluate per-frame image quality using the Fréchet Inception Distance (FID) \cite{heusel2018ganstrainedtimescaleupdate}, which measures the distributional similarity between generated frames and ground-truth frames. 
To assess temporal coherence, we use the Fréchet Video Distance (FVD) \cite{unterthiner2019accurategenerativemodelsvideo}, which measures both appearance and motion consistency by comparing distributions of video clips extracted from the generated and ground-truth sequences.

\paragraph{Efficiency.}
A central goal of \methodname{} is efficient video generation. 
We therefore report the \emph{generation time} required to synthesize each video, measured in wall-clock time on a single NVIDIA GH200 Superchip. The speed numbers reported for the experiments
on full video generation include both keyframes generation and the subsequent 3D
reconstruction and rendering. The numbers reported for the interpolation
baselines include only interpolation.

\begin{figure}[t]
    \centering
    \includegraphics[width=1.0\linewidth]{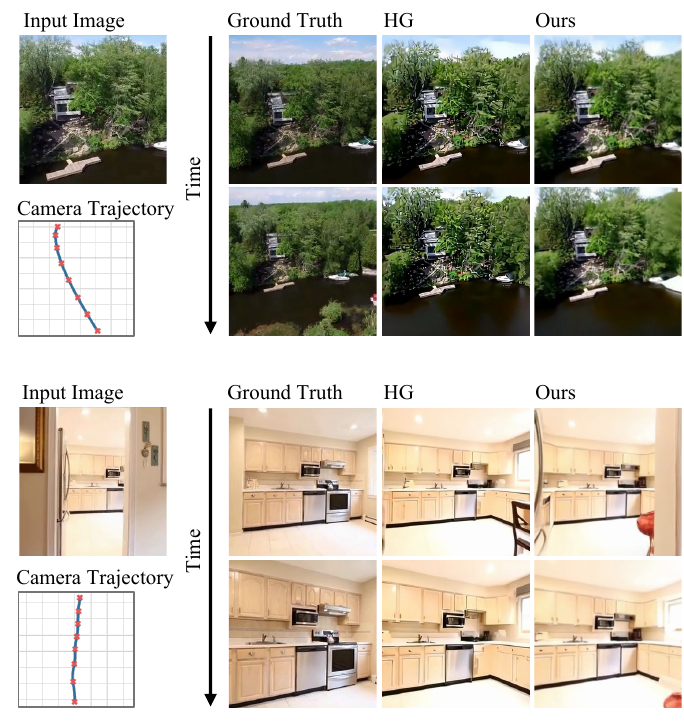}
    \caption{\textbf{Qualitative comparisons on RE10K.} Compared with HG~\cite{song2025historyguidedvideodiffusion}, our method does not have high-frequency artifacts and is significantly faster. }
    \label{fig:comparison_re10k}
\end{figure}

\begin{table*}[t]
\centering
\renewcommand{\arraystretch}{1.1}
\setlength{\tabcolsep}{5pt}
\begin{tabular}{l|cccc|cccc}

 &
\multicolumn{4}{c|}{{DL3DV (20s @ 30 fps, 600 frames)}} &
\multicolumn{4}{c}{{RE10K (20s @ 10 fps, 200 frames)}} \\
\cline{1-9}
 Method / Metric & FID $\downarrow$ & FVD $\downarrow$ & Time (s) $\downarrow$ & Speed-up $\uparrow$ &
   FID $\downarrow$ & FVD $\downarrow$ & Time (s) $\downarrow$ & Speed-up $\uparrow$ \\
\hline
HG~\cite{song2025historyguidedvideodiffusion} & 66.89 & 367.5 & 697.38 & 1$\times$ & 39.53 & 194.0 & 226.5 & 1$\times$ \\
Ours     & \textbf{60.90} & \textbf{335.5} & \textbf{16.21} & \textbf{43.02}$\times$ & \textbf{30.23} & \textbf{180.3} & \textbf{9.552} & \textbf{23.71}$\times$ \\
\hline
\end{tabular}
\caption{\textbf{Quantitative comparison on DL3DV and RE10K}. We achieve better results quantitatively at significantly higher speeds.}
\label{tab:comparison_combined}
\end{table*}

\subsection{Quantitative Results}

Table~\ref{tab:comparison_combined} summarizes the quantitative comparison on RE10K and DL3DV. 
Across both datasets, \methodname{} outperforms the History-Guided Video Diffusion (HG) baseline on both quality metrics, FID and FVD. 
This demonstrates that replacing dense diffusion with our keyframe diffusion and 3D rendering pipeline does not degrade visual quality, and in fact yields more consistent image and video generation.

The primary advantage of \methodname{} is its computational efficiency. 
We achieve more than 40$\times$ speed-up over HG on DL3DV, and more than 20$\times$ speed-up on RE10K. 
Notably, \methodname{} reaches real-time performance on DL3DV, requiring only 16.21 seconds on average to generate a 20-second 30~fps video at $256\times256$ resolution, corresponding to a generation framerate of 37.01 fps. In comparison, HG achieves a generation framerate of 0.86 fps. 
These results validate that explicit 3D reconstruction and rendering provide substantial efficiency gains without sacrificing fidelity.

\begin{table}[h]
\centering
\begin{tabular}{lcccc}
\hline
Method & FID $\downarrow$ & FVD $\downarrow$ & Time (s) $\downarrow$ & Speed-up $\uparrow$\\
\hline
Voyager~\cite{huang2025voyagerlongrangeworldconsistentvideo} & 91.75 & 808.0 & 332.0 & 1$\times$ \\
HG~\cite{song2025historyguidedvideodiffusion} & 62.78 & 497.8 & 116.0 & 2.86$\times$ \\
Ours & \textbf{61.18} & \textbf{492.8} & \textbf{13.62} & \textbf{24.38}$\times$ \\
\hline
\end{tabular}
\caption{\textbf{Quantitative comparisons on DL3DV (5 fps).} We achieve better or comparable quality, with significant speed-ups.}
\label{tab:comparisondl3dv5fps}
\end{table}

Table~\ref{tab:comparisondl3dv5fps} reports results on the 5~fps test set of DL3DV, 
where we compare against both the HG and the state-of-the-art Voyager model. 
Across all metrics, \methodname{} again achieves superior image quality (FID) and video quality (FVD), 
while being significantly faster than both baselines. 

\begin{figure}[t]
    \centering
    \includegraphics[width=1.0\linewidth]{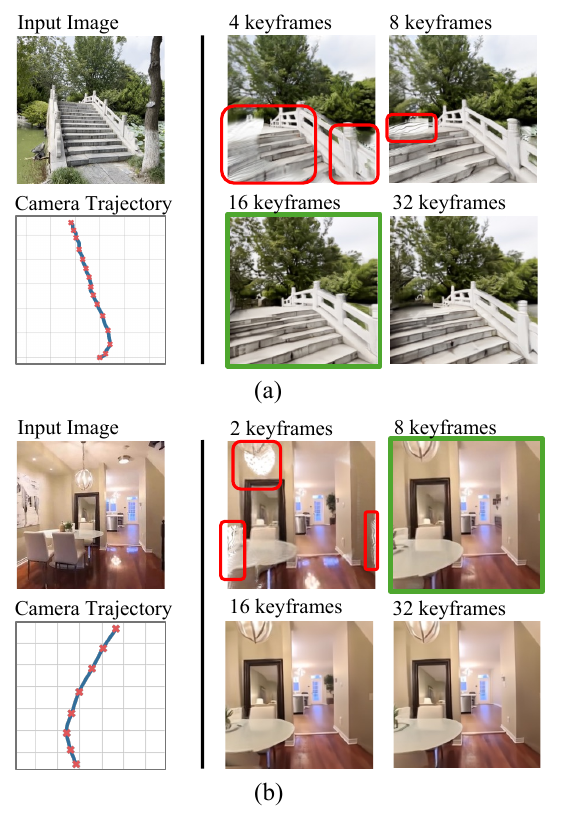}
    \vspace{-0.7cm}
    \caption{Too few keyframes lead to visible holes in the generated video (red boxes), while generating too many keyframes is significantly more expensive, without significant quality gains. \methodname{} selects the optimal number of keyframes (green) that strikes a good balance between completeness and efficiency.}
    \label{fig:ablation_keyframe_density}
\end{figure}

\subsection{Qualitative Results}

Figure~\ref{fig:comparison_dl3dv} shows qualitative comparisons on DL3DV against the HG and the Voyager model. 
Our method produces high-quality results under large viewpoint changes. 
Figure~\ref{fig:comparison_re10k} provides additional comparisons on RE10K, where \methodname{} again achieves high visual fidelity and consistent appearance across the trajectory. 
We refer readers to our website for full visualizations of the generated sequences.
Although the rendered videos from our 3D reconstruction pipeline may appear slightly smoother and contain fewer high-frequency details than the results from the diffusion-based baselines, they avoid the characteristic high-frequency artifacts produced by the HG model and maintain significantly better geometric stability than Voyager, exhibiting consistent structure and appearance across the entire trajectory.

\begin{table}[h]
\centering
\begin{tabular}{lcccc}
\hline
Method & FID $\downarrow$ & FVD $\downarrow$ & Time (s) $\downarrow$ & Speed-up $\uparrow$\\
\hline
FILM \cite{reda2022film} & \textbf{58.94} & 619.0 & 315.0 & 1$\times$ \\
RIFE \cite{huang2022rife} & 59.72 & 653.0 & 2.67 & 117.98$\times$ \\
Ours & 65.87 & \textbf{482.0} & \textbf{0.83} & \textbf{379.52$\times$} \\
\hline
\end{tabular}
\caption{\textbf{Comparison with 2D interpolation methods on DL3DV.} All methods are given the same sets of sparse keyframes $\sim$3s apart, generated by our keyframe generation model. The baselines have significant morphing effects, as reflected in the high FVD scores.
In addition, both baselines are significantly slower. Only interpolation time is reported here.}
\label{tab:2D3D}
\end{table}

\subsection{Ablation Studies}

\paragraph{3D vs.\ 2D Interpolation.}
A natural question is whether the dense video can be generated by applying a 2D interpolation method to the keyframes, rather than performing 3D reconstruction and rendering. 
Figure~\ref{fig:comparison_dl3dv} (rightmost two columns) and Table~\ref{tab:2D3D} compare our approach with two state-of-the-art 2D interpolation models, FILM~\cite{reda2022film} and RIFE~\cite{huang2022rife}. 
Our method outperforms both baselines in terms of FVD and avoids the morphing and warping artifacts commonly observed when interpolating across large viewpoint changes. 
Interestingly, \methodname{} is also \emph{faster} than the 2D interpolation baselines, since 3DGS rendering is highly efficient and scales well to long sequences. 
These results highlight the importance of explicit 3D reasoning for generating geometrically consistent videos.

\paragraph{Effect of Temporal Chunking.} 
\begin{table}[h]
\centering
\begin{tabular}{lcccc}
\hline
Method & FID $\downarrow$ & FVD $\downarrow$ & Time (s) $\downarrow$\\
\hline
HG~\cite{song2025historyguidedvideodiffusion} & 63.00 & 346.5 & 491.0\\
Ours (without chunking) & 62.84 & 357.5 & 13.52\\
Ours (with chunking) & \textbf{59.19} & \textbf{336.5} & \textbf{13.24}\\
\hline
\end{tabular}
\caption{\textbf{Evaluation of temporal chunking on DL3DV.} Both FID and FVD are improved with temporal chunking, while the computational time used is comparable. This experiment is conducted on generating 400-frame 30~fps videos.}
\label{tab:chunking}
\end{table}

We evaluate the quantitative performance of our method when generating long, high-frame-rate videos from the DL3DV test dataset, with or without dividing the keyframes into chunks and constructing separate 3D Gaussians. As shown in Table~\ref{tab:chunking}, both FID and FVD are improved when temporal chunking is used. Under the temporal chunking setting, the 3D reconstruction model can produce 3D scenes with higher consistency and reduced blurriness, and this is essential for rendering high-quality videos from the scene. Running the 3D reconstruction model multiple times also does not slow down the overall process. The model needs to produce the 3D scene from fewer keyframes each time, making each run faster than a single run over all keyframes together.

\paragraph{Keyframe Selection.} A key design choice of our method is that we train a model to predict the keyframe density given an input image and a camera trajectory. In Figure~\ref{fig:ablation_keyframe_density}, we can see that when too few keyframes are selected, the 3D scene is underdefined, and there will be multiple visible blank areas in the rendered video. However, when enough keyframes can define the scene along the given trajectory, adding more keyframes does not necessarily improve the quality and will add to computational redundancy. Our keyframe model chooses the optimal keyframe density that balances final video quality and computational cost.

\section{Discussion}
\paragraph{Static Scenes.}
Our method currently only applies to static scenes. 
We emphasize, however, that camera-conditioned generation of static videos is an important and well-studied problem in its own right, with numerous recent works focusing exclusively on this setting~\cite{huang2025voyagerlongrangeworldconsistentvideo, he2024cameractrl, li2025cameras}. 
Moreover, 4D reconstruction of dynamic scenes is progressing rapidly and has begun to benefit downstream tasks~\cite{zhang2024monst3r, wang2025continuous, wang2025shapemotion4dreconstruction, li2024megasamaccuratefastrobust, yao2025smallgsgaussiansplattingbasedcamera}. As these models mature, the core ideas behind \methodname{}, i.e., sparse view generation, adaptive keyframing, and 3D rendering, will directly be transferable to dynamic environments. 
We view our method as establishing a foundation for efficient generation that can be extended to full dynamic scenes in future work.

\paragraph{High-frequency Details.}
Our 3D-rendered videos may appear slightly smoother or less detailed at high frequencies compared to purely diffusion-based baselines. 
At the same time, they avoid the characteristic artifacts of diffusion models, such as noise amplification or view-dependent distortions, and maintain strong geometric consistency across challenging camera trajectories. 
As demonstrated in our experiments, this trade-off leads to quantitatively better FID/FVD scores and substantially faster generation. 
We also note that many practical applications, such as in embodied AI, do not require the highest-frequency details, but instead prioritize global coherence and structural stability. 
Finally, as 3D reconstruction models improve, we expect the visual fidelity of our results to increase correspondingly.

\section{Conclusion}

We presented a simple yet effective approach for efficient camera-conditioned video generation. 
By explicitly leveraging the inherent redundancy in video data, our method 
produces long, geometrically consistent videos at a small fraction of the computational cost of existing diffusion-based models. 
Our experiments demonstrate that this strategy not only yields dramatic speed-ups, but also maintains or improves visual quality relative to strong baselines. 
We believe that the core ideas behind \methodname{} will serve as a foundation for future work on efficient video synthesis.

\section{Acknowledgments}

We would like to thank Amani Kiruga for his help with processing datasets. We acknowledge the use of resources provided by the Isambard-AI National AI Research Resource (AIRR). Isambard-AI is operated by the University of Bristol and is funded by the UK Government’s Department for Science, Innovation and Technology (DSIT) via UK Research and Innovation; and the Science and Technology Facilities Council [ST/AIRR/I-A-I/1023]~\citep{isambard}.
{
    \small
    \bibliographystyle{ieeenat_fullname}
    \bibliography{main}
}

\appendix

\section{Method Details}
\label{sec:further method}
\subsection{Adaptive Keyframe Density Predictor}
\subsubsection{Model Architecture}
The model takes a dense camera trajectory and a reference image as input and outputs the number of sparse keyframes required. It comprises four components: a camera embedding module, a DINOv2 encoder, a transformer, and a final output MLP.

We embed the camera trajectory by converting the 3×3 rotation matrices into 4D quaternions, appending them to the translation vectors to form 7D vectors, and projecting them to have the same dimensionality as DINOv2 using a 2-layer MLP. We process the reference image with the DINOv2 encoder and retain only the global feature token, discarding all other image tokens. This token is appended to the sequence of camera tokens. The transformer, configured with 4 heads and 4 layers, processes this sequence and produces output tokens. The output tokens are passed through a 4-layer MLP to regress the final number of sparse keyframes.

\subsubsection{Training}

To stabilize the training, instead of regressing a single number as the keyframe density, we have the final MLP layers regress one number for each of the output tokens from the transformer and take the average of them as the final output of the keyframe density predictor. Additionally, as the keyframe counts in our curated dataset are in the range of 4 to 35, we find that scaling the final output by 0.1 regulates the output and improves the stability of the training. Specifically, the training loss of this model can be written as:
\begin{equation}
    \mathcal{L} = \mathbb{E}\!\left[\, (\bar{y} -  0.1 *  n_{gt})^{2} \right]
\end{equation}
where $\bar{y}$ is the average of the output vector of the MLP, and $n_{gt}$ is the ground truth keyframe number. We optimize the network over the entire dataset, which is collected using the full video sequences in each dataset. The dataset is collected using camera trajectories from 10-fps RE10K videos and 5-fps DL3DV videos. We use 5-fps videos on DL3DV rather than 30-fps, as the keyframe sequence naturally will have a much lower frame rate than the final rendered video. In the training split, the average length of the full video sequences is 134.94 frames at 10 fps for the RE10K dataset and 339.48 frames at 5 fps for the DL3DV dataset.

We train the model using AdamW with a learning rate of 1e-4 and a batch size of 128. Training was performed on a single NVIDIA GH200 Superchip for 5 hours.

\subsubsection{Uniform Keyframe Sampling}
While directly predicting the exact keyframe indices would be ideal, this objective is highly unstable to train due to the inherent ambiguity of the problem. Instead, we regress the total number of sparse keyframes required and uniformly distribute them along the camera trajectory, which yields better results.

\subsection{Keyframe Generation Model}
Our keyframe generation model is a history-guided video diffusion model (HG model)~\cite{song2025historyguidedvideodiffusion}. We train a camera-controlled, diffusion-forcing transformer (DFoT) on a sparse keyframe dataset with camera pose annotations.

\subsubsection{Training}
Conventional video diffusion models are trained to denoise an entire video from the same noise level. In contrast, the diffusion-forcing architecture assigns an independent noise level to each frame at training time and trains the model to denoise all the frames within the context window from different noise levels together, minimizing the noise prediction loss:
\begin{equation}
    \mathcal{L} = \mathbb{E}
    \left[ \left\| \epsilon_{\mathcal{T}} - \epsilon_{\theta}\!\left( x_{\mathcal{T}}^{k_{\mathcal{T}}},\, k_{\mathcal{T}} \right) \right\|^{2} \right],
\end{equation}

where $\tau$ is the set of frame indices in a video, $x_{\mathcal{T}}^{k_{\mathcal{T}}}$ denotes noisy frames, $k_{\mathcal{T}}$ denotes the associated noise levels, and $\epsilon_{\mathcal{T}}$ is the noise added to the frames. $\epsilon_\theta()$ is a neural network that predicts the added noise, and $\theta$ denotes the model parameters to be optimized. The parameters are optimized to minimize the noise prediction loss across all frames and noise levels.

When training our keyframe generation model, we start from the checkpoint trained on the RE10K dataset, provided by \cite{song2025historyguidedvideodiffusion}, and finetune the model for the DL3DV dataset. As stated in Section 3.2.2, we first train the model with consecutive frames and then progressively increase the gap between adjacent frames to approximately 4 seconds, matching the keyframe density predicted by our keyframe density predictor. We trained the model with a batch size of 8 and a learning rate of 5e-5 on 8 NVIDIA GH200 Superchips for 6 days. 

\subsubsection{Inference}
As the diffusion-forcing models are trained to denoise frames with independent noise levels, they can use either the input image or the previously generated frames as part of the condition by giving the conditional frames to the model with zero noise level. This model characteristic enables autoregressive sampling and infinite video generation. These models can also accept guidance at arbitrary positions, enabling interpolation. 

For all our experiments, we use the vanilla history guidance, which performs classifier-free guidance with a preset image condition length, and adhere to the default parameter settings as stated in the HG GitHub repository\footnote{https://github.com/kwsong0113/diffusion-forcing-transformer}. Camera control is provided by encoding the camera poses into ray maps, following the HG paper. Our model uses a context window of 8 frames. When generating more than 8 keyframes, we first generate 8 keyframes that span the entire trajectory using only the provided input image as conditioning. The remaining keyframes are then generated using the same model, with the nearest already generated keyframes as conditioning.

\subsection{3D Reconstruction Model}
As specified in Section 3.3, we use AnySplat~\cite{jiang2025anysplatfeedforward3dgaussian} as our 3D reconstruction model to obtain the 3D Gaussians scene representation from the generated keyframes. AnySplat takes uncalibrated images as input and uses a transformer-based geometry encoder to extract the latent representations of the images. It then utilizes three decoder heads to predict the Gaussian parameters of the scene, the depth map, and the camera poses associated with each input image. Finally, the Gaussian parameters are voxelized into per-voxel 3D Gaussians using a Differentiable Voxelization module. The model is trained using an RGB loss between the input images and the renderings from the predicted 3D Gaussians, along with a geometry loss that uses the corresponding estimates from the pretrained VGGT model~\cite{wang2025vggt} to supervise the predicted camera poses and depth maps. The RGB loss consists of a mean-squared error term and a Learned Perceptual Image Patch Similarity (LPIPS)-based loss. The geometry loss is the mean-squared error between the depth maps and camera encoding estimated by the two models. 
In our pipeline, we use the pretrained model of AnySplat. This model is trained on several datasets, including DL3DV, but not RE10K. We feed AnySplat our generated keyframes to obtain the 3D Gaussians as well as the predicted keyframe camera poses for subsequent processing.

\subsection{Camera Pose Alignment for Rendering}
\subsubsection{Aligning the Camera Trajectory with the 3D Reconstruction}
\label{sec:align_gt_anysplat}
The input camera trajectory and 3D Gaussians predicted by AnySplat may differ in scale, rotation, and translation. We compute an affine transformation to align the input trajectory with the 3D reconstruction’s coordinate system. We do so by first estimating the least squares affine transformation between AnySplat's predicted camera poses for the keyframes and the camera poses for the keyframes from the full input camera trajectory.
\begin{equation}
S^{*}
= \arg\min_{S}
\sum_{k \in K}
\left\| S \circ \Phi_k^{\mathrm{in}} - \Phi_k^{\mathrm{as}} \right\|^{2},
\end{equation}
Then, we apply the estimated transformation to the whole input camera trajectory. 

\begin{equation}
\widehat{\Phi}_i^{\mathrm{as}} = S^{*} \circ \Phi_i^{\mathrm{in}} .
\end{equation}

Here, $K$ is the set of keyframes, $\Phi_k^{\mathrm{in}}$ denotes the keyframe camera poses in the full input trajectory, and $\Phi_k^{\mathrm{as}}$ denotes the keyframe camera poses predicted by AnySplat for the 3D Gaussians. $\Phi_i^{\mathrm{in}}$ is the full input trajectory with $i$ being the frame index. $S$ is the affine transformation, and $S \circ \Phi$ denotes applying the affine transformation $S$ to the camera pose $\Phi$. Finally, we render the whole video using the transformed full camera trajectory $\widehat{\Phi}_i^{\mathrm{as}}$.

\subsubsection{Aligning the Camera Trajectory under Temporal Chunking}

Similarly, when we divide the video into temporal chunks and reconstruct each chunk independently, each chunk may have its own coordinate system. To produce a seamless video, we include a shared keyframe between consecutive chunks and align the dense input camera trajectory to each chunk individually. We further compute and apply a transformation 
to the initially aligned camera trajectory to ensure the transformed shared keyframe poses at each chunk are identical to the shared keyframe poses predicted by AnySplat. This step is essential for a smooth transition across the chunk boundary. The full video is rendered chunk by chunk, and duplicate renders of the overlapping keyframes are discarded.

\section{Experimental Details}
\label{sec:further exp}
\subsection{Datasets}
The RE10K dataset used in our experiments was provided by History-Guided Video Diffusion~\cite{song2025historyguidedvideodiffusion} and was obtained from \url{https://huggingface.co/kiwhansong/DFoT/tree/main/datasets}
. All videos are at 256×256 image resolution and at 10 fps. 200 videos that have at least 200 frames from the RE10K test split are chosen, and the first 200 frames of each video are used in our experiments with RE10K.

The DL3DV evaluation set is downloaded from \url{https://huggingface.co/datasets/DL3DV/DL3DV-Evaluation}. We resize all videos to a height of 256 pixels and center-crop them to obtain square frames, ensuring compatibility with the History-Guided Video Diffusion (HG) baseline and our method. Although the videos in this dataset are at a higher fps, the provided camera poses are only at 5 fps. We interpolate the camera poses to be 30 fps so that camera-controlled video generation at a higher frame rate can be evaluated using this dataset. The rotation is interpolated using slerp, and the translation is interpolated linearly. The primary evaluation against the HG model is conducted on generating 20-second 30-fps videos. For comparisons with a wider range of baseline methods, the subsampled 5-fps videos are used.

\subsection{Detailed Experiment Setup}
Following the default setting of HG~\cite{song2025historyguidedvideodiffusion} for generating 200-frame 10 fps videos from the RE10K dataset, we compute the baseline results of the HG model on the RE10K dataset by first generating 12 frames uniformly spanning the full trajectory (keyframes) with the input image as conditioning, and then generating the remaining frames using the nearest two keyframes as conditioning. 
We maintain the same initial keyframe density in time for the HG baseline for all experimental setups, which is 12 keyframes in 20 seconds. The original HG model was not trained on the DL3DV dataset. For a fair comparison, we train a separate HG model on the DL3DV dataset as the baseline model. This model is trained with the same progressive strategy until the gap between the adjacent frames in the training videos is 2 seconds. This training setting is comparable to that of the HG model for the RE10K dataset described in the original paper~\cite{song2025historyguidedvideodiffusion}.

When generating 200-frame 10-fps videos on RE10K and 100-frame 5-fps videos on DL3DV with our pipeline, we use the camera poses of the respective target length as the inputs to the keyframe density prediction model, as the model supports variable-length inputs. When generating 30-fps videos on the DL3DV dataset, we first temporally subsample the input camera trajectory to 5 fps and feed the subsampled camera trajectory to the keyframe density predictor, ensuring that the inputs remain within the training distribution. After the 3D reconstruction is obtained from the predicted keyframes, the final video is rendered using the full 30-fps camera trajectory.

\subsection{Comparisons with Voyager}
Originally, the Voyager model~\cite{huang2025voyagerlongrangeworldconsistentvideo} was trained on rectangular images. However, we evaluate models on square videos to be compatible with the HG baseline and our method. To accommodate this, when running the experiments with the Voyager model on the DL3DV dataset, we use the full rectangular frames from the dataset for generation and perform the evaluation on the center-cropped square videos. This gives slightly higher performance than performing inference with the square image directly. 

It is worth mentioning that Voyager claims to be able to generate arbitrarily long videos with autoregressive sampling and World Caching with Point Culling. However, this part of the method is not found in their official implementation\footnote{https://github.com/Tencent-Hunyuan/HunyuanWorld-Voyager}. Thus, we only run the experiment with the Voyager model on the setting of generating 100-frame videos at 5~fps. We show qualitative results on our supplemental webpage, and quantitative results in Table 2 of the main paper.

\subsection{Comparisons with 2D interpolation methods}
We compare our interpolation method using 3D reconstruction with two 2D video interpolation methods, RIFE~\cite{huang2022rife} and FILM~\cite{reda2022film}. We first generate keyframes using our keyframe generation model at a fixed keyframe density. Then, we use the same set of keyframes for the 2D interpolation methods and our method to interpolate the full video.

Our method shows superior results both qualitatively and quantitatively, as shown in Table 3 (main paper). Neither of the 2D methods considers camera control when interpolating between frames, resulting in naively morphing when the camera control is complex.

\section{Videos of the Same Scene with Different Camera Trajectories}
After generating the 3D Gaussians and a video with our method, we can render new videos with different camera trajectories, even those that are out-of-distribution with respect to the trajectories observed during training, in seconds. The baseline diffusion model, on the other hand, would need to rerun the generation process, which requires hundreds of seconds. Please refer to the supplemental webpage for the qualitative results.

\end{document}